# Low Precision Neural Networks using Subband Decomposition

Sek Chai, Aswin Raghavan, David Zhang, Mohamed Amer, Tim Shields

SRI International, Princeton, NJ

*Abstract*— **Large-scale deep neural networks (DNN) have been successfully used in a number of tasks from image recognition to natural language processing. They are trained using large training sets on large models, making them computationally and memory intensive. As such, there is much interest in research development for faster training and test time. In this paper, we present a unique approach using lower precision weights for more efficient and faster training phase. We separate imagery into different frequency bands (e.g. with different information content) such that the neural net can better learn using less bits. We present this approach as a complement existing methods such as pruning network connections and encoding learning weights. We show results where this approach supports more stable learning with 2-4X reduction in precision with 17X reduction in DNN parameters.**

## I. INTRODUCTION

It is well understood in the deep learning community that to capture the wide spectrum of low, mid and high-level representations for deep semantic understanding of complex patterns, networks with many layers, nodes, and with high local and global connectivity are needed. The success of recent deep learning algorithms (e.g., in speech, vision and natural language) comes in part from the ability to train much larger models on much larger datasets than was previously possible [1-3]. One fundamental challenge is that as the dataset size sequences increases, the degree of non-linearity that is involved increases exponentially. This makes it hard to capture complex spatiotemporal dependencies thereby impacting their efficiency, training time and performance.

Over the last decade, DNN parameter sizes continue to grow dramatically. In 1998, the LeNet-5 Convolution Neural Net (CNN) [4] uses 1M parameters to classify handwritten digits. The AlexNet CNN [1] uses 60M parameters to win the ImageNet image classification competition in 2012. To give a perspective, for automatic recognition of 1000 categories of objects in ImageNet dataset, this DNN has seven layers, supporting 65K neurons and 630M synaptic connections. We estimate 3000 Gflops/s assuming 512x512 images at 100 Gflops/frame. More recently, new DNNs such as Deepface uses 120M parameters for human face verification [5], and there are even networks with 10B parameters [6]. Han et al. [7] estimates 12.8W just for DRAM access for DNN with 1B connections, and reiterates that this power and memory size exceeds the budget for a typical mobile device.

Our goal is to directly address the exponential growth in memory sizes for large DNNs by exploring new approaches that accommodate low precision weights. These new approaches should not take a binary or carte blanche application simple truncation, rounding or recasting of floating points to lower precision integers because the resulting DNN gradient estimates would have very high variance and the overall learning will not converge. Instead, our focus is on a scalable approach for all layers of the network with minimal variance and precision loss in algorithmic performance.

Our particular approach is inspired by image compression using wavelet decomposition [8]. An image is first decomposed into multiple frequency bands such that the transient elements of a data signal can be represented by a smaller amount of information. High-frequency components that are beyond the acuity of the human visual system can be removed for lossy compression and reduction in file size.

In our low precision DNN, we adopt a similar notion that an image can be decomposed into different frequency bands and that the opportunities to trim precision comes after DNN learning. We argue that this method can better guide the DNN to learn specific features specifically, and from this more guided learning, we can better afford to trim precision without loss in algorithmic performance. In this paper, we show the efficacy of this approach where we separate images into high and low frequency bands, and allow DNN to learn edge and textures separately. Much like the lossy compression, we could justify the precision reduction based on removal of learn representation that is not needed for the image recognition task.

This paper is organized as follows: in Section 2, we present comparisons against related research in low precision DNN. In Section 3, we define a DNN architecture that highlights our approach as well as the baseline architecture for comparison. In Section 4, we describe our simulation setup for MNIST and CIFAR-10 datasets, and we provide early results and associated analysis that evaluates our low precision approach. Finally in Section 5, we present our conclusions and discuss our future work in this space.

## II. RELATED RESEARCH

DNN are typically over-parameterized in that there is significant redundancy in the learning model. On the one hand, the sparsity of the learnt representations offer DNN the high algorithmic performance, but at the same time, it is easy to arrive at a DNN that is a waste in both computation and memory. Recent research efforts have started to address DNN size and complexity. Some of these efforts are motivated from the need to reduce training time (e.g. from weeks of processing time on large server clusters), and also from the desire to migrate DNN to small low power mobile processors. The key goal in all of these efforts is to find the balance where the DNN sparsity (and therefore robustness and algorithm performance) is maintained without over-parameterized the model.

One basic approach to low precision deals with approximation and quantization techniques. Vanhoucke et al. [9] converted 32-bit floating point activations to 8-bit integer fixed point implementations. Gong et al. [10] compressed DNN weights using vector quantization. Gupta et al. [11] explore stochastic rounding techniques to implement DNN from 32-bit floating point to 16-bit fixed point integers on FPGA. Another approach is focused on training the DNNs with defined target weights [12-15]. Courbariaux et al. [16] shows a training method using binary weights to avoid the forward and backward propagation multiplications.

Network connection pruning has been a subject of very recent works. Han et al. [7] proposes a multi-step training method where neural connections are removed based on a weighted threshold. Similarly, Kung et. al [17] studied the energy and accuracy tradeoffs by removing neural connections based on error propagation.

Even though these approaches share the same objective, the methods are quite different. Our approach on subband decomposition is orthogonal and can be used together to obtain additional improvements. Our approach using decomposition relies on pre-processing the input data into different subbands. This process is different from data augmentation, where images are rotated, mirrored, contrast adjusted, in order to generate sufficient variation of the original images for training. Data decomposition is a process that separates the energy or information content, and can work in conjunction to data augmentation for additional training benefits.

## III. APPROACH

In this section, we provide details on our subband decomposition approach to low precision DNN. Our main goal is to study its characteristics in order to delineate the efficacy. To accomplish this, we perform analysis on a baseline DNN without the different variants and methods that researchers have used to incrementally gain in algorithmic performance.

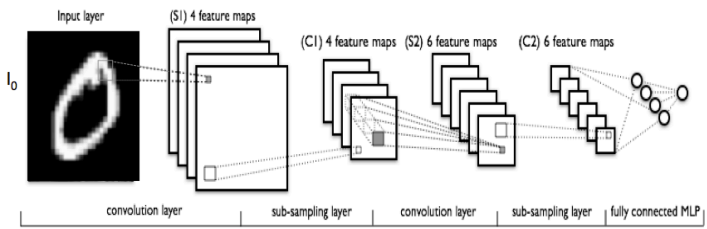

Figure 1. Baseline architecture for comparison with "original" or unaltered input image Io.

We choose the LeNet-5 CNN, as shown in Figure 1, with five layers comprising of two convolutional layers with 5x5 filters and ReLU activation function. A pooling/subsampling layer, implementing the max pooling function over non-overlapping pooling windows of size 2x2, follows each convolutional layer. The second pooling layer feeds into a fully connected layer consisting of another ReLU layer, which is then connected into a 10-way softmax output layer. This DNN model is well studied and should allow for good reference point for our study. In our study, we label results from this architecture as "original" with input image Io.

Figure 2 illustrates our proposed DNN architecture that incorporates the decomposed input images. For this study, we show two subbands, high and low, generated by Laplacian and Gaussian filters [18]. When decomposed, they represent edge and texture respectively, and we labeled decomposed images as Lo and G1. Much like the pyramid structure in [18], we choose a subsampled representation of the Gaussian at half resolution.

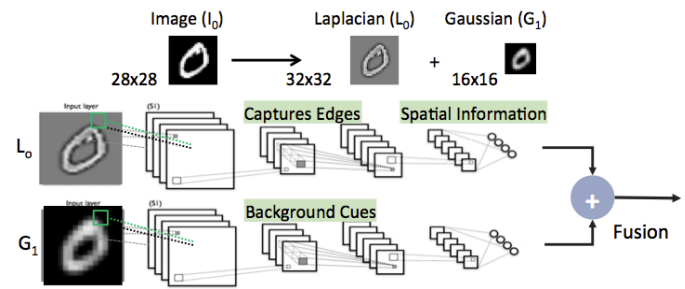

Figure 2. Proposed DNN architecture for two subbands, Lo and G1, representing edge and texture information.

For each subband, we provide a DNN to learn the spatial and spectral features for that particular frequency band. We have chosen the same LeNet-5 CNN architecture for simplicity in comparison. We provide a fusion stage where we combine results from each subband. For this study, we choose a simple equation with equi-weighted average of softmax output (see equation EQ1 below). This fusion stage can be a neural additional layer that is parameterized, but it is not considered in this study at this time.

$$S = 0.5*(s_1(x) + s_2(y)), \quad s_1(x), s_2(x) \text{ in } [0,1]^{10} \text{ and } ||s(x)||_1 = 1 \quad (EQ1)$$

During training, our method seeks to maintain the overall network size as the original baseline design. We achieve this using stochastic rounding [11] to trim the weights. Stochastic rounding is an unbiased rounding scheme with desirable property that the expected rounding error is zero. This is especially important to preserve the gradient information, statically over many training epochs. First we define the range of weight coefficients based on the target bit precision. Then we update the weights of each layer with SGD (Stochastic Gradient Descent) in back-propagation. We use a default learning rate of 0.1, but we do explore the effect of learning rate on this network in the next section.

## IV. RESULTS AND EVALUATION

For our study, we use both MNIST and CIFAR-10 data to cover both edge specific and natural/color images. The MNIST database contains black and white handwritten digits, normalized to 20x20 pixel size. There are 60,000 training images and 10,000 testing images. The CIFAR-10 dataset consists of 60000 32x32 color images in 10 classes, with 6000 images per class. There are 50000 training images and 10000 test images. Each of these images were prefiltered (Laplacian and Gaussian) and fed into the proposed DNN architecture. The baseline architecture uses only the original image (unaltered).

Our evaluation methodology includes algorithmic performance comparison of the proposed trained DNN against the baseline network architecture. We vary different algorithmic parameters to characterize their impact on the DNN to recognize objects in the 10 classes. This approach is a standard process to compare neural networks. Our results are reported as test or validation error, indicating how well the network can recognize objects in the dataset. A lower error value is desired. In this study, we do not perform any data augmentation so as not to influence the result with orthogonal approaches. We note that our results should be compared against the baseline and not against any competitive ranking for these benchmarks.

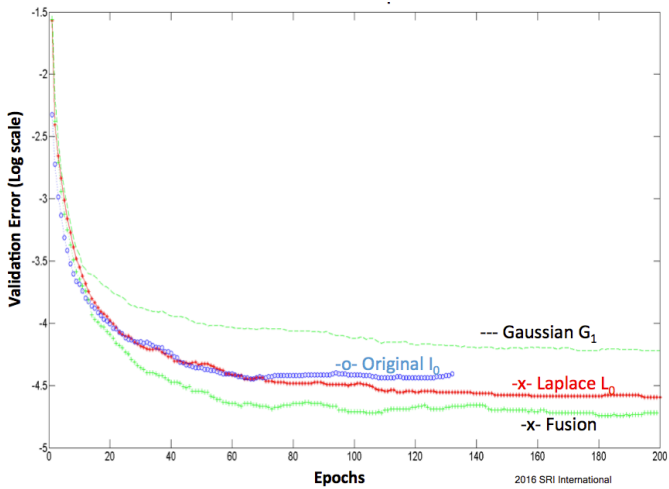

Figure 3. DNN performance for MNIST dataset, showing lower validation error for fused Lo+G1 result over baseline.

Figure 3 illustrates the performance of the proposed DNN architecture against the baseline. We show that the learning rate of the baseline DNN, using original Io image, converges at a level between the Lo and G1 results. We interpret this as follows. Because the MNIST dataset are digits, the Lo network is able to better capture more salient feature for digit classification. G1 network performs poorer because its input data are more texture oriented. It is interesting that the fusion of Lo and G1 results produce results that is 16% better.

The results in Figure 3 assume full 32bit floating point for all networks. That is, we did not trim the weights for Lo and G1 yet. This is done intentionally to illustrate that the subband decomposition approach can result in better overall algorithmic performance. We attribute this from the subband decomposition's ability to separate features from "noise" from a learning perspective. That is, we are able to direct tune the learning for each subband, allowing the overall learning at a more optimal point.

| Stochastic Rounding after final epoch | | | | |
|---|---|---|---|---|
| Weight bits | 32bit | 16bit | 8bit | 4bit |
| Original | 1.13 | 1.22 | 1.14 | 4.45 |
| GBlur | 1.38 | 1.34 | 1.35 | **5.24** |
| Laplace | 1.02 | 0.96 | 1.03 | 6.03 |
| Fusion | **0.89** | **0.91** | **0.92** | 6.03 |

| Stochastic Rounding after every epoch | | | | |
|---|---|---|---|---|
| Weight bits | 32bit | 16bit | 8bit | 4bit |
| Original | 1.13 | 1.13 | 1.11 | 1.32 |
| GBlur | 1.38 | 1.41 | 1.36 | 1.38 |
| Laplace | 1.02 | 1.00 | 0.91 | 1.66 |
| Fusion | **0.95** | **0.89** | **0.89** | **1.13** |

Figure 4. DNN performance for MNIST dataset, with varying bit precisions.

Figure 4 shows the DNN performance for the MNIST dataset for different bit precision. The left table shows result when stochastic rounding is performed at the end of the training sequence. The right table shows result when the stochastic rounding is performed after each epoch. It is obvious that fine grain guidance helps reduce the learning gradient and therefore the results are better (on the right table).

With respect to precision, we note that the subband approach can reduce the bit precision by a half. The 32-bit baseline architecture can be scaled down to 8-bits before performance is affected at 4-bits. Similarly, the fusion exhibits the same behavior at 4-bits. However, at 4-bit precision, the fusion result is comparable or better than the 16-bit and 8-bit baseline.

We note that the baseline network takes similar amount of time to train compared to each subband in the proposed DNN. There is some overhead in computing the Lo and G1 subbands, but they are negligible when compared to the processing involved in training. We are still characterizing and optimizing the computation using CuDNN on NVIDIA GPUs.

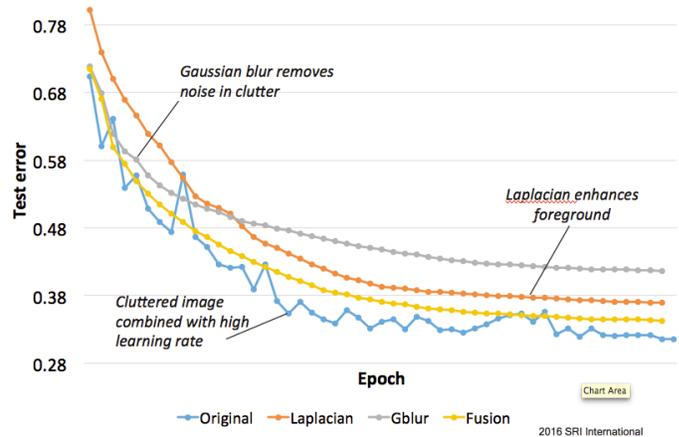

Figure 5. DNN performance for CIFAR-10 dataset, showing comparable test error for fused result over baseline

Figure 5 shows the results using the CIFAR-10 dataset. Here, we note that the Lo and G1 results have an interesting crossover in performance during training. We attribute this to the nature of CIFAR-10 image as some of the images can be simply classified with color patches (e.g. airplane classes are easily distinguished with blue sky). The fusion results remain at par with the baseline results as they both converge similarly. It is important to highlight the oscillation of the results for the baseline in comparison to the fused results. We suspect that this is due to the variation of features in the original Io images. Just as in the MNIST data, we attribute this to subband approach to separate features from "noise" from a learning perspective.

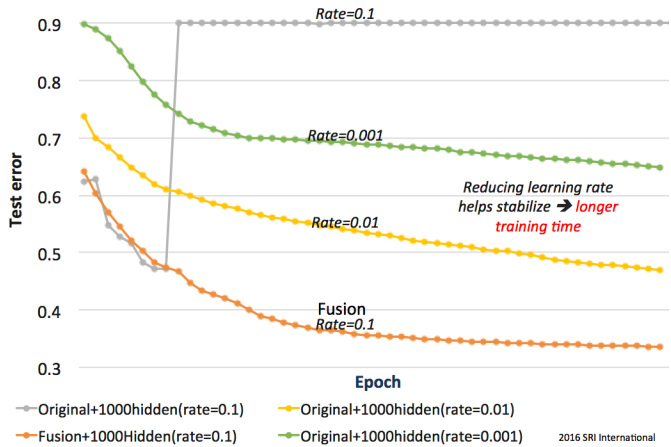

Figure 6. Effect of learning rate, showing the stability of fusion results at higher learning rate.

To better illustrate the learning stability, we explore the effects on different learning rate. As shown in Figure 6, the original baseline DNN can converge in a stable manner at lower learning rate (0.01 and 0.001). However, at a rate of 0.1, the DNN may be at a tethering at a stability point. In comparison, the fusion results shows stable learning at the rate of 0.1.

In this experiment, we increase the hidden layers to 1000 neurons to dramatically highlight this behavior. The anticipated influence is that the subband approach can afford a faster training time because the original baseline can be prone to stability issues with higher learning rates.

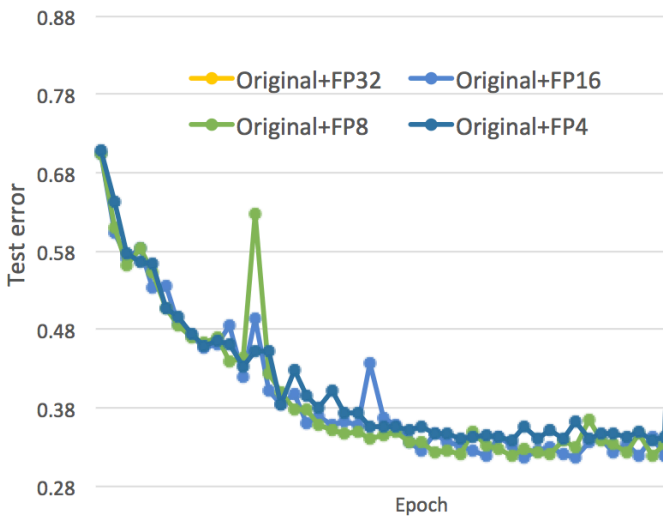

Figure 7. Baseline DNN performance for CIFAR-10 dataset, with varying bit precisions.

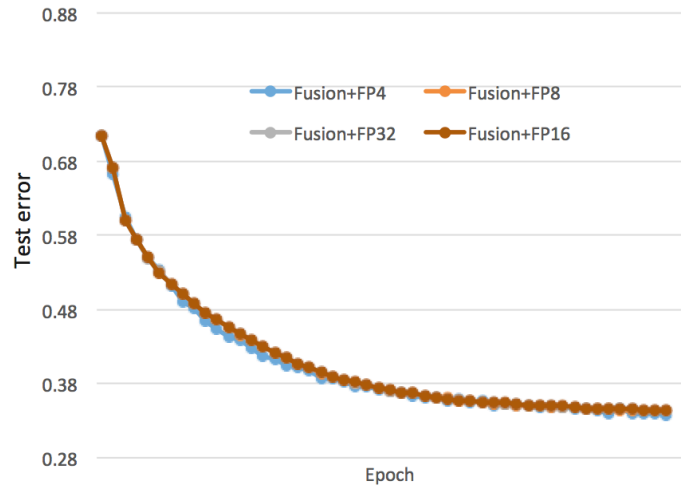

Figure 8. Subband DNN performance for CIFAR-10 dataset, with varying bit precisions.

Figure 7 and Figure 8 shows the DNN performance for the CIFAR-10 dataset for different bit precision. We show that both baseline and subband approach both converges to a similar level. We already noted that the oscillating results in baseline results are due to the higher learning rate. It is important to note that algorithmic performance is similar for precision from 32-bit to 4-bit for both baseline and subband approach.

However, we reiterate that learning stability is an important factor, and that the subband approach supports both lower precision and stable learning, at similar learning rates. As shown in Figure 9, comparing 4-bit fusion with baseline 32-bit, we demonstrate this desired characteristic against more erratic baseline result. With respect to the fused subband network, the largest size network (32-bit floating point, with 1000 hidden units) uses 11.7M parameters, while the smallest size network (4-bit floating point with 50 hidden units) uses only 0.68M parameters. This amounts to a reduction in of 17X parameters while maintaining stable learning and comparable performance.

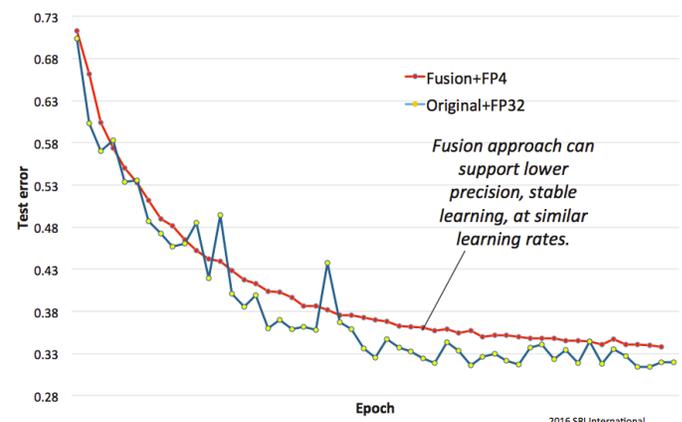

Figure 9. Desired stable learning provided by subband decomposition (fusion) even at lower precision (4-bit versus 32-bit) weights.

## V.  Conclusions and Future Work

In this paper, we explore the notion of decomposing an image into different subbands in order to arrive at a more optimal DNN learning. This characteristic helps in achieving better classification results (per the MNIST result), or more stable learning (per the CIFAR-10 result).

We also present the subband approach as an alternative means to design lower precision DNN. The basic premise is that if each subband is more "optimal" from the learning perspective, we can better able to quantize and approximate the learning weights. We show that our approach works orthogonally with stochastic rounding approach to lower precision. We anticipate that we will be able to combine other approaches such as network pruning to further improve the performance. The costs associated with our proposed approach involve processing the subbands during the training phase.

In this brief study, we provide a simple fusion approach to combine the subband results. For future work, we plan to study a more principled approach with an additional network layer where the weighted precision can be learned. We presented a very early implantation to show the basic principles of subband DNN. We anticipate we will find more interesting results with larger datasets with higher resolution images such as ImageNet. For those datasets, we may have to migrate to a deeper network that might show different behaviors with respect to performance and learning stability.

Benefits of the data size reduction when using smaller bit precision is best observed during the inference mode. For example, the much reduced data size helps reduces memory bandwidth requirements, which allows very fast and power efficient approaches. In smartphone processors, the smaller data size may allow much of the DNN weights to fit in processor cache. The lower precision approaches also support faster training time because the process of using smaller precision acts as a regularizer by inducing bias and decreasing variance of the network parameters.

Going forward, there is much more we can do as future work to extend the notion of image decomposition. For example, we can learn in different color space such as HSV and YCrCb. We may also consider Gabor filters instead of Laplacian/Gaussian filters, which are more biologically inspired. The non-linearity in some of these color transformations from RGB may not be easily learned, and the prefiltering step may help the DNN learn important features to improve its algorithmic performance. We would like to explore hardware implementation on FPGA hardware to delineate the memory and power efficiency gains offered by the subband decomposition approach.

## VI.  Acknowledgements

This research was developed with funding from the Defense Advanced Research Projects Agency (DARPA) and the Air Force Research Laboratory (AFRL). The views, opinions and/or findings expressed are those of the authors and should not be interpreted as representing the official views or policies of the Department of Defense or the U.S. Government.